%% file: main.tex
\documentclass[10pt,twocolumn,letterpaper]{article}

\usepackage[pagenumbers]{cvpr} 

\input{preamble}

\definecolor{cvprblue}{rgb}{0.21,0.49,0.74}
\usepackage[pagebackref,breaklinks,colorlinks,allcolors=cvprblue]{hyperref}

\def\paperID{*****} 
\def\confName{CVPR}
\def\confYear{2026}

\title{SAM3-DMS: Decoupled Memory Selection for Multi-target Video Segmentation\\ of SAM3\vspace{-0.03in}}

\author{
Ruiqi Shen\textsuperscript{1} \qquad Chang Liu\textsuperscript{2,\Letter} \qquad Henghui Ding\textsuperscript{1,\Letter} \\
\textsuperscript{1}Fudan University \qquad \textsuperscript{2}Shanghai University of Finance and Economics \\[0.25em]
\normalsize{\url{https://github.com/FudanCVL/SAM3-DMS}}\vspace{-0.4in}
}

\begin{document}
\maketitle

\begin{strip}
    \centering
    \includegraphics[width=0.96\textwidth]{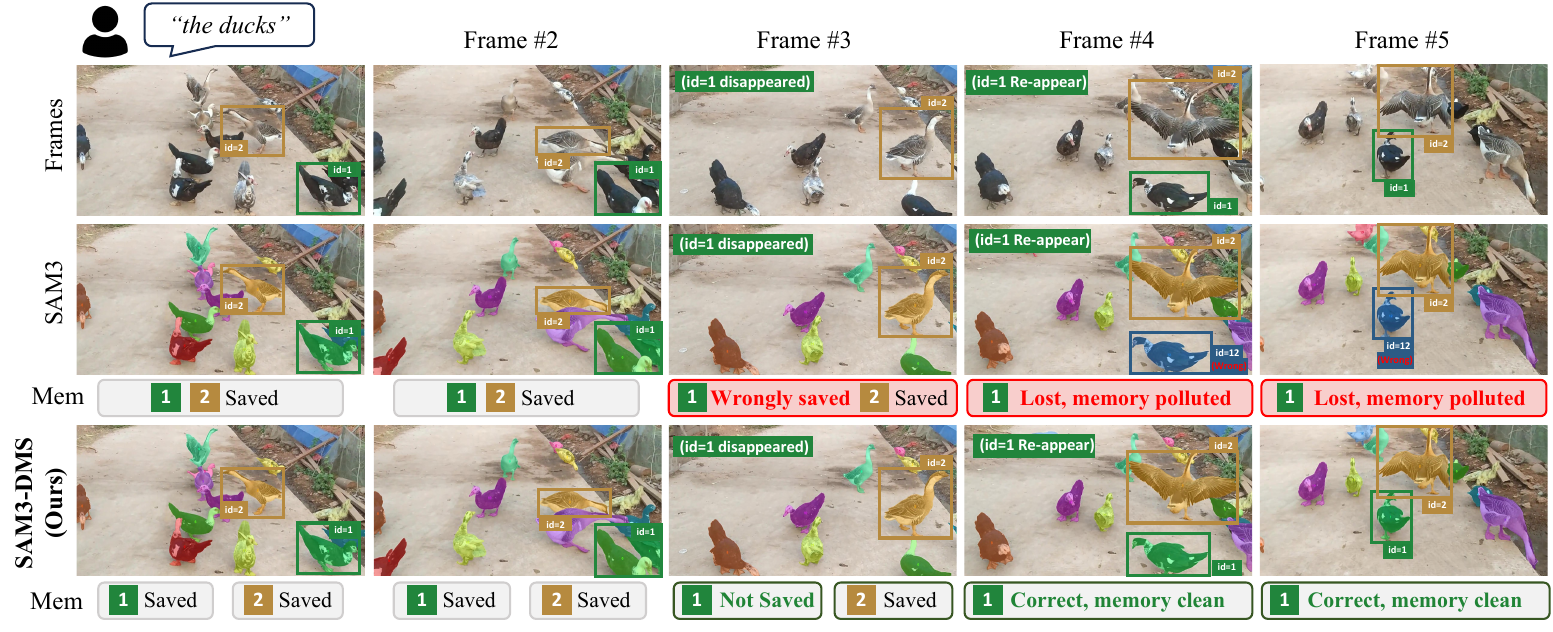}
    \vspace{0.02in}
    \captionof{figure}{
    SAM3 \textit{vs.} SAM3-DMS (Ours), in simultaneous multi-object video segmentation. In Frame \#3, SAM3 updates the memories for all objects together based on the overall status, saving Object \colorbox{obj1}{\textcolor{white}{1}} into memory even its mask is blank, causing identity drifts. In contrast, we separately update memory for each object by its own, keeping consistent identity tracking. 
    }
    \vspace{0.03in}
    \label{fig:teaser}
\end{strip}
\renewcommand{\thefootnote}{\fnsymbol{footnote}}
\footnotetext[0]{${\textrm{\Letter}}$ Corresponding to: {hhding@fudan.edu.cn}, {liuchang@sufe.edu.cn}}

\input{sec/0_abstract}    
\input{sec/1_intro}
\input{sec/2_formatting}

\clearpage
{
    \small
    \bibliographystyle{ieeenat_fullname}
    \bibliography{main}
}


\end{document}

%% file: preamble.tex
\usepackage{placeins}
\usepackage{cuted}
\usepackage{caption}

\usepackage{graphicx}
\usepackage{multirow}
\usepackage{amsmath}
\usepackage{subcaption}

\usepackage{tikz}
\usepackage{xcolor}
\usepackage{subcaption}
\usepackage{makecell}
\usepackage{tablefootnote}

\usepackage{marvosym}

\definecolor{obj1}{HTML}{23762E}
\definecolor{obj2}{HTML}{7C1819}

%% file: sec/0_abstract.tex
\begin{abstract}


Segment Anything 3 (SAM3) has established a powerful foundation that robustly detects, segments, and tracks specified targets in videos. However, in its original implementation, its group-level collective memory selection is suboptimal for complex multi-object scenarios, as it employs a synchronized decision across all concurrent targets conditioned on their average performance, often overlooking individual reliability. To this end, we propose SAM3-DMS, a {training-free} decoupled strategy that utilizes fine-grained memory selection on individual objects. Experiments demonstrate that our approach achieves robust identity preservation and tracking stability. Notably, our advantage becomes more pronounced with increased target density, establishing a solid foundation for simultaneous multi-target video segmentation in the wild.
\end{abstract}

%% file: sec/1_intro.tex
\section{Introduction}
\label{sec:intro}

The Segment Anything Model (SAM) family has revolutionized pixel-wise visual perception, with SAM1 \cite{kirillov2023segment} and SAM2 \cite{ravi2024sam} introducing Promptable Visual Segmentation (PVS) for images and videos, while SAM3 \cite{carion2025sam} advances Promptable Concept Segmentation (PCS) by detecting and tracking instances of high-level semantic concepts, marking a significant step toward high-level video grounding.

Built upon SAM2, SAM3 employs a memory bank to preserve target context across frames. To handle complex scenarios such as temporary disappearance and re-entry, SAM3 implements a ``memory selection'' strategy which conditionally updates the memory bank by thresholding the prediction confidence, memorizing only the reliable features. However, in simultaneous multi-target video segmentation scenarios, this selection strategy remains limited to coarse frame-level decision. Specifically, by aggregating confidence scores of distinct targets into a group-level average, SAM3 enforces a synchronized update decision across all concurrent targets driven by their collective performance, rather than the reliability of each individual.

While this approach suffices for single-object video segmentation, it may become problematic in real-world videos where the scene contains multiple targets, each with distinct and complex patterns, such as PCS cases. \cref{fig:teaser} shows a demonstration where Object \#1 disappears from the scene in Frame \#3. However, SAM3 overlooks this absence as the high confidence of other objects (\eg, \#2)  elevates the group-level average score. Consequently, a blank mask is encoded into and polluted Object \#1's memory, causing identity drift at its subsequent re-entry.

To maintain temporal identity consistency of individual targets while adhering to the simultaneous paradigm of tracking and segmenting all targets at once, we propose SAM3-DMS, a simple yet effective training-free approach. Firstly, to address the group-level memory selection issue, we employ an decoupled policy where the selection is decided on each object independently, rather than the average of all concurrent targets. Secondly, the decision is subject to the target's self-assessment. In each frame, each target derives a confidence score by combining its segmentation score with the overall visibility status, which serves as a criterion to determine whether or not to update its own corresponding memory. By eliminating distractions from other unrelated targets in the group, this fine-grained memory selection strategy ensures robust multi-target video segmentation in challenging scenarios involving occlusion, distractors, and target re-entries, as illustrated in \cref{fig:teaser}.

Furthermore, we perform comprehensive evaluations for our decoupled memory selection strategy on both PCS and PVS tasks across seven benchmarks and in-the-wild videos under the simultaneous multi-target video segmentation setting, which demonstrate the superiority of our approach, notably revealing that the performance gap widens as target density increases. In summary, this work enhances SAM3's multi-target robustness driven by fine-grained memory control. By decoupling memory maintenance conditioned on individual reliability, SAM3-DMS effectively mitigates identity drift and secures temporal consistency in challenging open-world scenarios.

\section{Related Work}
\textbf{Memory-based VOS.} \ Memory-based networks have become the dominant paradigm in Video Object Segmentation (VOS) \cite{pont20172017}, with the Space-Time Memory (STM) \cite{oh2019video} serving as the foundational framework. Recent advances include STCN \cite{cheng2021rethinking} for optimized attention computation, XMem \cite{cheng2022xmem} for decoupling long- and short-term memory, RMNet \cite{xie2021efficient} with optical flow assistance, and Cutie \cite{cheng2024putting} for object-level guidance. SAM2 \cite{ravi2024sam} and SAM3 \cite{carion2025sam} have emerged as the state-of-the-art VOS methods. Trained on millions of samples, they demonstrate exceptional video segmentation performance. Recent works improve SAM2 by introducing memory trees \cite{ding2025sam2long}, motion modeling \cite{yang2024samurai}, and memory partitioning \cite{videnovic2025distractor}. \\

\noindent \textbf{Promptable Concept Detection and Segmentation} is formulated as visual grounding, which aims to localize specific targets within an image or video based on user prompts in natural language. Existing approaches mainly fall into two paradigms: the first builds directly upon specialist detection or segmentation architectures, incorporating auxiliary language encoders for precise alignment \cite{li2022grounded, liu2024grounding, ren2024grounded}. The other adopts Large Vision-Language Models (LVLMs) for grounding, either treating grounding as an autoregressive sequence generation task by directly outputting coordinates of bounding boxes \cite{chen2023shikra, chen2024expanding, wang2025spacevllm, li2025llava,MeViS,MeViSv2} or utilizing special tokens to prompt external decoders for fine-grained mask prediction \cite{lai2024lisa, bai2024one, gong2025devil, yuan2025sa2va}. Despite their success, these approaches struggle to maintain consistent identities in dynamic environments where targets exhibit frequent entries, exits, and reappearances. 
SAM3 \cite{carion2025sam} bridges this gap by unifying detection, segmentation, and tracking to ensure robust performance in such dynamic scenarios.

%% file: sec/2_formatting.tex
\section{Method}

\begin{figure*}[t]
    \centering
    \includegraphics[width=1.0\textwidth]{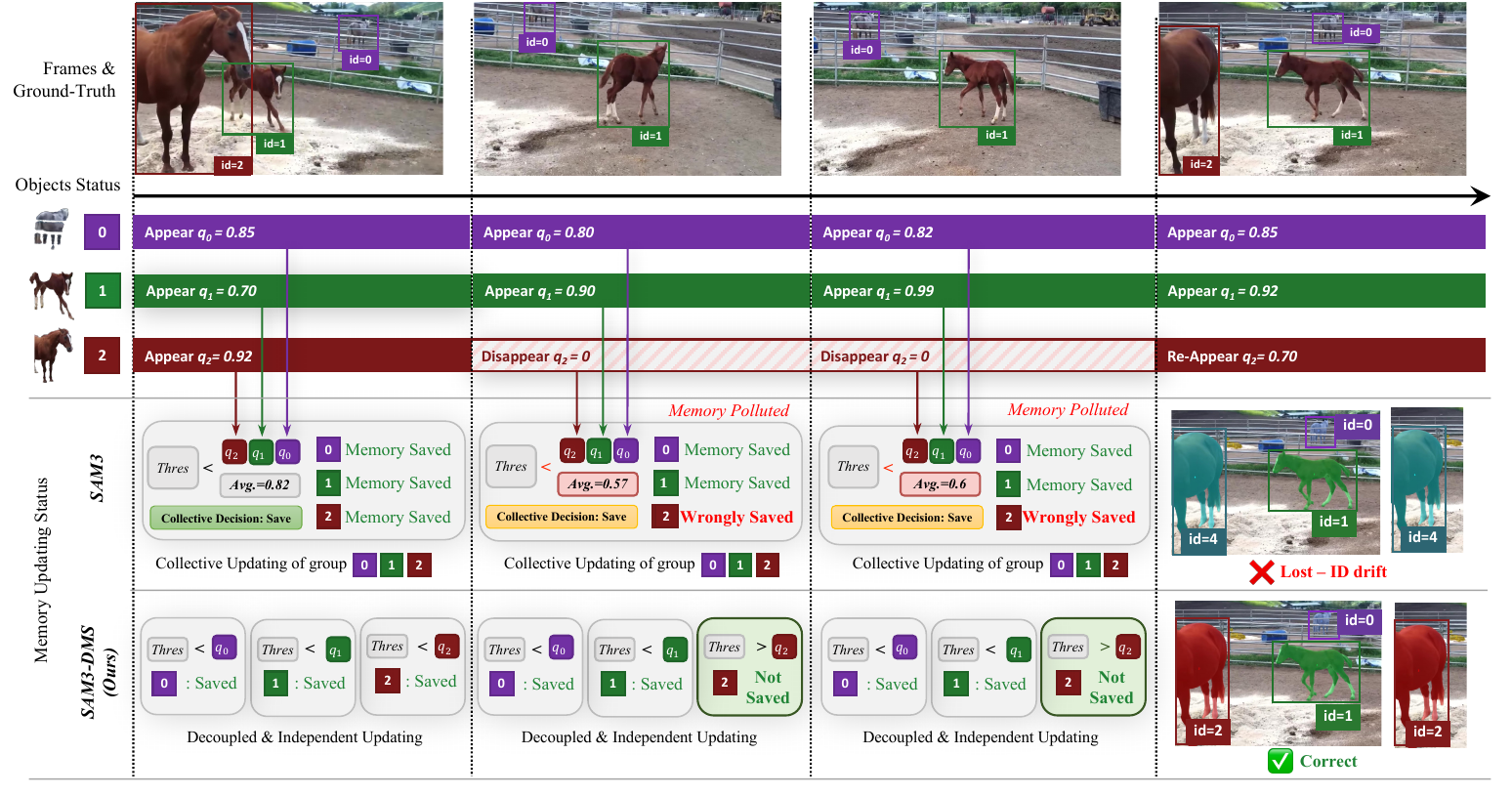}
    \caption{
        Overview of the Decoupled Memory Selection (DMS) mechanism. In SAM3, the memory status of the group is determined \textit{collectively} by the average score ($Avg$), causing out-of-view objects to be ``Wrongly Saved'' when other group members remain visible. Our SAM3-DMS evaluates each target's status \textit{independently}. This decoupling prevents corrupted features from entering the memory bank, leading to the correct identity preservation of Object \colorbox{obj2}{\textcolor{white}{2}} seen in the final frame.
    }
    \label{fig:fig2}
\end{figure*}

\subsection{Preliminaries: Memory Selection of SAM3} 
\label{sec:Preliminaries}
We focus on simultaneous multi-target video segmentation, in which all objects are inputted to the model at once, rather than inferring for each target at a time. Let \( \mathcal{O} = \{ o_{1,2,\dots ,N} \} \) denote targets instantiated in a same initial frame. For each target $o_i$ at time \( t \), SAM3 predicts a mask \( M_{i,t} \) and its corresponding segmentation score (or \textit{``query's score''}) \( q_{i,t} \) reflecting the object's status such as segmentation confidence and individual disappearance status, along with a frame-level presence score \( p_t \) demonstrating the overall visibility of all objects in the frame.

Following the design of SAM2 \cite{ravi2024sam}, SAM3 does not contain shared object-level contexts and essentially maintains a dedicated memory bank $\mathcal{M}_i$ for each target $o_i$ \cite{carion2025sam}. Despite this, 
SAM3 implements a unified memory selection policy across all objects $\mathcal{O}$, where all \( N \) memory banks are either updated together or not at all. Specifically, it calculates an aggregated frame-level confidence score as follows:
\begin{equation}
    S_t = (\frac{1}{N} \sum_{i} q_{i,t}) \cdot p_t, \quad \forall i \in \{ 1, \dots, N \}.
\end{equation}
For target $o_i$, let $\mathbf{f}_{i,t} = \Phi(I_t, M_{i,t})$ denote the features encoded by the memory encoder $\Phi$ and $\mathbf{p}_{i,t}$ be the associated object pointer. The update of its memory bank $\mathcal{M}_i$ at frame $t$ is conditioned on the frame-level average $S_t$:

\begin{equation}
    \mathcal{M}_i \leftarrow 
    \begin{cases} 
      \mathcal{M}_i \cup \{ (\mathbf{f}_{i,t}, \mathbf{p}_{i,t}) \}, & \text{if } S_t > \tau \\
      \mathcal{M}_i, & \text{otherwise}
    \end{cases}
    \label{eq:coupled_decision}
\end{equation}
where $\tau$ is a pre-defined confidence threshold. Relying exclusively on this group average, this strategy allows high-performing salient objects to mask the unreliability of uncertain or disappearing objects. As illustrated in \cref{fig:fig2}, due to the existence of two high-score objects (\#0, \#1), the overall confidence score exceeds the threshold and causing the model to save all objects' features, including the disappeared object (\#2).
This pollutes its memory bank and leads to the following identity drift when the disappeared object re-entries the scene, misidentifying it as a new object (\#4). 

Notably, for PCS tasks with language prompts where new objects may emerge mid-video, objects that are initialized at the same timestamp are grouped together and all subject to this memory selection policy.

\subsection{SAM3-DMS: Decoupled Memory Selection}
\label{sec:decoupled_memory}

To mitigate group-level interference and ensure target-level discriminability, we propose the Decoupled Memory Selection (DMS) strategy. Instead of relying on the frame-level \( S_t \), we perform memory selection in an instance-wise manner. This ensures the memory maintenance strictly follows the individual tracking status of its corresponding target. Specifically, the confidence \( S_{i,t} \) is computed for each target \( o_i \) by combining its own segmentation score \( q_{i,t} \) with the presence score \( p_t \) as follows:
\begin{equation}
    S_{i,t} = q_{i,t} \cdot p_t, \quad \forall i \in \{ 1, \dots, N \},
\end{equation}
which is then thresholded for the update decision of its corresponding memory bank \( \mathcal{M}_i \):
\begin{equation}
\mathcal{M}_i \leftarrow
\begin{cases}
\mathcal{M}_i \cup \{ (\mathbf{f}_{i,t}, \mathbf{p}_{i,t}) \}, & \text{if } S_{i,t} > \tau \\
\mathcal{M}_i, & \text{otherwise}
\end{cases}
\label{eq:decoupled_decision}
\end{equation}
thereby ensuring that memory maintenance is driven exclusively by the quality of each individual target. As shown in \cref{fig:fig2}, our decoupled approach recognizes the Object \#2's temporary disappearance and excludes blank frames from its memory. This prevents memory pollution, ensuring clean representations for seamless re-identification upon its re-entry in the last frame.

Fundamentally, this decoupled paradigm enables individualized memory update frequencies, allowing each target’s representation to evolve independently according to its specific tracking status. Moreover, since the underlying memory allocation and architecture remain identical to that of SAM3, these improvements are achieved with virtually no additional GPU memory overhead.

\begin{table*}[!ht]
\centering
\caption{Quantitative comparisons of PCS on the SA-Co/VEval benchmark. \textsuperscript{*}: Reproduced using official implementation.}
\label{tab:pcs_a}
\renewcommand{\arraystretch}{1}
\setlength{\tabcolsep}{11pt}
\begin{tabular}{c l l l l l}
\toprule
Splits & Methods & cgF1 & pHOTA & pDetA & pAssA \\
\midrule
\multirow{3}{*}{\makecell[c]{SA-V \\ val}}
  & SAM3 & 29.3 & 60.7 & 44.7 & 83.2 \\
  & SAM3\textsuperscript{*} & 29.2 & 60.68 & 44.63 & 83.32 \\
  & \textbf{SAM3-DMS}\textsuperscript{*} & \textbf{29.4} (\textcolor{red}{$\uparrow$0.2}) & \textbf{60.92} (\textcolor{red}{$\uparrow$0.24}) & \textbf{44.89} (\textcolor{red}{$\uparrow$0.26}) & \textbf{83.50} (\textcolor{red}{$\uparrow$0.18}) \\
\midrule
\multirow{3}{*}{\makecell[c]{SA-V \\ test}}
  & SAM3 & 30.3 & 58.0 & 40.9 & 83.4 \\
  & SAM3\textsuperscript{*} & 30.1 & 57.80 & 40.64 & 83.30 \\
  & \textbf{SAM3-DMS}\textsuperscript{*} & \textbf{30.3} (\textcolor{red}{$\uparrow$0.2}) & \textbf{57.97} (\textcolor{red}{$\uparrow$0.17}) & \textbf{40.83} (\textcolor{red}{$\uparrow$0.19}) & \textbf{83.41} (\textcolor{red}{$\uparrow$0.11}) \\
\midrule
\multirow{3}{*}{\makecell[c]{YT-Temporal-1B \\ val}}
  & SAM3 & 50.2 & 70.5 & 60.5 & 82.7 \\
  & SAM3\textsuperscript{*} & 49.8 & 70.03 & 59.64 & 82.88 \\
  & \textbf{SAM3-DMS}\textsuperscript{*} & \textbf{50.3} (\textcolor{red}{$\uparrow$0.5}) & \textbf{70.28} (\textcolor{red}{$\uparrow$0.25}) & \textbf{59.98} (\textcolor{red}{$\uparrow$0.34}) & \textbf{82.99} (\textcolor{red}{$\uparrow$0.11}) \\
\midrule
\multirow{3}{*}{\makecell[c]{YT-Temporal-1B \\ test}}
  & SAM3 & 50.8 & 69.9 & 60.2 & 81.7 \\
  & SAM3\textsuperscript{*} & 49.9 & 69.17 & 59.24 & 81.39 \\
  & \textbf{SAM3-DMS}\textsuperscript{*} & \textbf{51.0} (\textcolor{red}{$\uparrow$1.1}) & \textbf{69.88} (\textcolor{red}{$\uparrow$0.71}) & \textbf{59.91} (\textcolor{red}{$\uparrow$0.67}) & \textbf{82.15} (\textcolor{red}{$\uparrow$0.76}) \\
\midrule
\multirow{3}{*}{\makecell[c]{SmartGlasses \\ val}}
  & SAM3 & 33.5 & 60.2 & 46.2 & 79.3 \\
  & SAM3\textsuperscript{*} & 33.2 & 60.08 & 46.08 & 79.02 \\
  & \textbf{SAM3-DMS}\textsuperscript{*} & \textbf{33.6} (\textcolor{red}{$\uparrow$0.4}) & \textbf{60.29} (\textcolor{red}{$\uparrow$0.21}) & \textbf{46.17} (\textcolor{red}{$\uparrow$0.09}) & \textbf{79.41} (\textcolor{red}{$\uparrow$0.39}) \\
\midrule
\multirow{3}{*}{\makecell[c]{SmartGlasses \\ test}}
  & SAM3 & 36.4 & 63.6 & 50.0 & 81.5 \\
  & SAM3\textsuperscript{*} & 36.1 & 63.44 & 49.90 & 81.28 \\
  & \textbf{SAM3-DMS}\textsuperscript{*} & \textbf{36.5} (\textcolor{red}{$\uparrow$0.4}) & \textbf{63.73} (\textcolor{red}{$\uparrow$0.29}) & \textbf{50.11} (\textcolor{red}{$\uparrow$0.21}) & \textbf{81.68} (\textcolor{red}{$\uparrow$0.40}) \\
\bottomrule
\end{tabular}
\vspace{0.1in}
\end{table*}

\begin{table*}[!ht]
\centering
\caption{Ablation study on SA-Co/VEval across varying target densities. YT: YT-Temporal-1B, SG: SmartGlasses.}
\renewcommand{\arraystretch}{1}
\label{tab:pcs_b}
\begin{tabular}{c l ll ll ll}
\toprule
\multirow{2}{*}{Splits} & \multirow{2}{*}{Methods} & \multicolumn{2}{c}{$\geq 3$ targets} & \multicolumn{2}{c}{$\geq 8$ targets} & \multicolumn{2}{c}{$\geq 10$ targets} \\
\cmidrule(lr){3-4} \cmidrule(lr){5-6} \cmidrule(lr){7-8}
& & cgF1 & pHOTA & cgF1 & pHOTA & cgF1 & pHOTA \\
\midrule
\multirow{2}{*}{\makecell[c]{SA-V \\ test}}
  & SAM3\textsuperscript{} & 39.78 & 53.13 & 41.29 & 56.30 & 41.00 & 53.74 \\
  & \textbf{SAM3-DMS}\textsuperscript{} & \textbf{39.98} (\textcolor{red}{$\uparrow$0.20}) & \textbf{53.37} (\textcolor{red}{$\uparrow$0.24}) & \textbf{42.02} (\textcolor{red}{$\uparrow$0.73}) & \textbf{56.83} (\textcolor{red}{$\uparrow$0.53}) & \textbf{42.37} (\textcolor{red}{$\uparrow$1.37}) & \textbf{54.70} (\textcolor{red}{$\uparrow$0.96}) \\
\midrule
\multirow{2}{*}{\makecell[c]{YT \\ test}}
  & SAM3\textsuperscript{} & 55.55 & 67.73 & 55.82 & 67.99 & 57.71 & 69.64 \\
  & \textbf{SAM3-DMS}\textsuperscript{} & \textbf{56.66} (\textcolor{red}{$\uparrow$1.11}) & \textbf{68.37} (\textcolor{red}{$\uparrow$0.64}) & \textbf{57.36} (\textcolor{red}{$\uparrow$1.54}) & \textbf{68.88} (\textcolor{red}{$\uparrow$0.89}) & \textbf{59.50} (\textcolor{red}{$\uparrow$1.79}) & \textbf{70.54} (\textcolor{red}{$\uparrow$0.90}) \\
\midrule
\multirow{2}{*}{\makecell[c]{SG \\ test}}
  & SAM3\textsuperscript{} & 48.29 & 63.48 & 43.71 & 61.26 & 46.29 & 65.27 \\
  & \textbf{SAM3-DMS}\textsuperscript{} & \textbf{48.80} (\textcolor{red}{$\uparrow$0.51}) & \textbf{63.75} (\textcolor{red}{$\uparrow$0.27}) & \textbf{44.46} (\textcolor{red}{$\uparrow$0.75}) & \textbf{61.84} (\textcolor{red}{$\uparrow$0.58}) & \textbf{46.81} (\textcolor{red}{$\uparrow$0.52}) & \textbf{65.65} (\textcolor{red}{$\uparrow$0.38}) \\
\bottomrule
\end{tabular}
\end{table*}

\begin{table*}
\centering
\caption{Quantitative comparisons of PCS on other public benchmarks. \textsuperscript{*}: Reproduced using official implementation.}
\label{tab:pcs_combined}
\renewcommand{\arraystretch}{1}
\setlength{\tabcolsep}{13.5pt}
\label{tab:pcs_c}
\begin{tabular}{c l l l l l}
\toprule
\multirow{2}{*}{Datasets} & \multirow{2}{*}{Methods} & Overall & $\geq 3$ targets & $\geq 5$ targets & $\geq 7$ targets \\
\cmidrule(lr){3-3} \cmidrule(lr){4-4} \cmidrule(lr){5-5} \cmidrule(lr){6-6}
& & $\text{mAP}$ & $\text{mAP}$ & $\text{mAP}$ & $\text{mAP}$ \\
\midrule
\multirow{2}{*}{\makecell[c]{YTVIS19 \\ val}}
  & SAM3\textsuperscript{*} & 58.38 & 51.13 & 48.32 & 73.35 \\
  & \textbf{SAM3-DMS}\textsuperscript{*} & \textbf{58.41} (\textcolor{red}{$\uparrow$0.03}) & \textbf{51.22} (\textcolor{red}{$\uparrow$0.09}) & \textbf{48.86} (\textcolor{red}{$\uparrow$0.54}) & \textbf{76.46} (\textcolor{red}{$\uparrow$3.11}) \\
\midrule
\multirow{3}{*}{\makecell[c]{YTVIS21 \\ val}}
  & SAM3 & 57.4 & - & - & - \\
  & SAM3\textsuperscript{*} & 56.85 & 53.10 & 50.94 & 47.39 \\
  & \textbf{SAM3-DMS}\textsuperscript{*} & \textbf{57.03} (\textcolor{red}{$\uparrow$0.18}) & \textbf{53.36} (\textcolor{red}{$\uparrow$0.26}) & \textbf{51.60} (\textcolor{red}{$\uparrow$0.66}) & \textbf{50.05} (\textcolor{red}{$\uparrow$2.66}) \\
\midrule
\multirow{3}{*}{\makecell[c]{OVIS \\ val}}
  & SAM3 & 60.5 & - & - & - \\
  & SAM3\textsuperscript{*} & 61.18 & 61.07 & 58.27 & 54.69 \\
  & \textbf{SAM3-DMS}\textsuperscript{*} & \textbf{62.26} (\textcolor{red}{$\uparrow$1.08}) & \textbf{62.13} (\textcolor{red}{$\uparrow$1.06}) & \textbf{59.94} (\textcolor{red}{$\uparrow$1.67}) & \textbf{56.19} (\textcolor{red}{$\uparrow$1.50}) \\
\midrule
\multicolumn{2}{c}{} & TETA & HOTA & DetA & AssA \\
\midrule
\multirow{3}{*}{\makecell[c]{BDD100K \\ val}}
  & SAM3 & 47.2 & - & - & - \\
  & SAM3\textsuperscript{*} & 49.49 & 40.45 & 31.26 & 55.13 \\
  & \textbf{SAM3-DMS}\textsuperscript{*} & \textbf{49.71} (\textcolor{red}{$\uparrow$0.22}) & \textbf{40.75} (\textcolor{red}{$\uparrow$0.30}) & \textbf{31.50} (\textcolor{red}{$\uparrow$0.24}) & \textbf{55.52} (\textcolor{red}{$\uparrow$0.39}) \\
\bottomrule
\end{tabular}
\end{table*}

\begin{table*}[!ht]
\centering
\captionsetup{width=\textwidth}
\caption{Quantitative comparisons of PVS on different benchmarks. \textsuperscript{*}: Reproduced using official implementation.}
\label{tab:PVS}
\renewcommand{\arraystretch}{1}
\begin{tabular}{p{1.9cm} p{1.9cm} p{1.55cm} p{0.6cm} p{0.6cm} | p{1.55cm} p{0.6cm} p{0.6cm} | p{1.55cm} p{0.6cm} p{0.6cm}}
\toprule
Methods & Mode & \multicolumn{3}{c}{SA-V val} & \multicolumn{3}{c}{SA-V test} & \multicolumn{3}{c}{MOSEv2 val} \\
& & $\mathcal{J\&F}$ & $\mathcal{J}$ & $\mathcal{F}$ & $\mathcal{J\&F}$ & $\mathcal{J}$ & $\mathcal{F}$ & $\mathcal{J}\&\dot{\mathcal{F}}$ & $\mathcal{J}$ & $\dot{\mathcal{F}}$ \\
\midrule
SAM3 & {One-by-one} & 83.5 & - & - & 84.4 & - & - & 60.3 & - & - \\ 
SAM3$^{*}$ & {One-by-one} & 83.3 & 79.4 & 87.2 & 84.3 & 80.4 & 88.3 & 60.3 & 57.9 & 62.7 \\
\midrule
SAM3 & {Simultaneous} & 81.2 & 77.4 & 84.9 & 81.3 & 77.4 & 85.2 & 60.0 & 57.6 & 62.4 \\
\textbf{SAM3-DMS} & {Simultaneous} & \textbf{83.3} (\textcolor{red}{$\uparrow$2.1}) & \textbf{79.4} & \textbf{87.2} & \textbf{84.3} (\textcolor{red}{$\uparrow$3.0}) & \textbf{80.4} & \textbf{88.3} & \textbf{60.3} (\textcolor{red}{$\uparrow$0.3}) & \textbf{57.9} & \textbf{62.7} \\
\bottomrule
\end{tabular}
\vspace{0.05in}
\end{table*}

\section{Experiments}

\subsection{Implementation Details}
Building upon SAM3, SAM3-DMS is training-free, fully preserving the original model's PCS and PVS capabilities. All experiments were conducted based on the official implementation of SAM3, on a single NVIDIA RTX A6000 GPU with 48GB of memory.

\subsection{Benchmarks and Metrics}

\textbf{PCS. } Following~\cite{carion2025sam}, we evaluate SAM3-DMS on the SA-Co/VEval benchmark containing 10.3K video-NP pairs, and report the cgF1, pHOTA, pDetA, and pAssA metrics. To further assess the generalizability of the model, we evaluate on extra public benchmarks, including YTVIS19 \cite{yang2019video}, YTVIS21 \cite{yang20213rd}, OVIS \cite{qi2022occluded}, and BDD100K \cite{yu2020bdd100k}, following the same evaluation metrics and protocols as~\cite{carion2025sam}, where category names are used as conceptual prompts.

\noindent \textbf{PVS. } We evaluate SAM3-DMS on SA-V (val/test)~\cite{ravi2024sam} and MOSEv2 (val)~\cite{ding2025mosev2,MOSE}, covering rich multi-target and occlusion scenarios. Following SAM3, we report the $\mathcal{J}\&\mathcal{F}$ metric for SA-V and $\mathcal{J}\&\dot{\mathcal{F}}$ metric for MOSEv2.

\subsection{Quantitative Experiments}
\textbf{PCS.} We firstly perform the evaluation on the SA-Co/VEval benchmark. As shown in Table \ref{tab:pcs_a}, SAM3-DMS consistently outperforms SAM3 on all metrics and datasets. Notably, as demonstrated in Table \ref{tab:pcs_b}, the performance gap becomes more significant with increasing target density. Specifically, as the scenes become extremely complex and contain more than 10 targets, our performance surpasses the original SAM3 by 1.79 cgF1 on the YT-Temporal-1B test set and 1.37 on the SA-V test set. Moreover, a similar trend is observed on other public benchmarks, as in Table \ref{tab:pcs_c}. For example, on the YTVIS21 validation set, the improvement of mAP scales from 0.18 (overall) to 0.66 ($\ge 5$ targets), ultimately reaching 2.66 ($\ge 7$ targets). These results demonstrate the effectiveness of our proposed DMS strategy in preserving identity for multi-target segmentation.

\noindent\textbf{PVS.} We additionally benchmark SAM3-DMS for simultaneous multi-target PVS. As shown in Table \ref{tab:PVS}, compared with the upper bound where each target is independently inputted and inferred by the network one-by-one, the shared memory selection used in the original SAM3 suffers from severe performance degradation. Our approach effectively bridges this gap, achieving gains of 2.1 and 3.0 $\mathcal{J}\&\mathcal{F}$ on SA-V val and test splits, and achieves almost no performance loss on both SA-V and MOSEv2 compared with the upper bound. Note in PCS, one-by-one inference is not applicable due to the dynamic emergence of new objects.

\begin{figure*}[!t]
  \centering
  \begin{subfigure}{\textwidth}
    \centering
    \includegraphics[width=\textwidth]{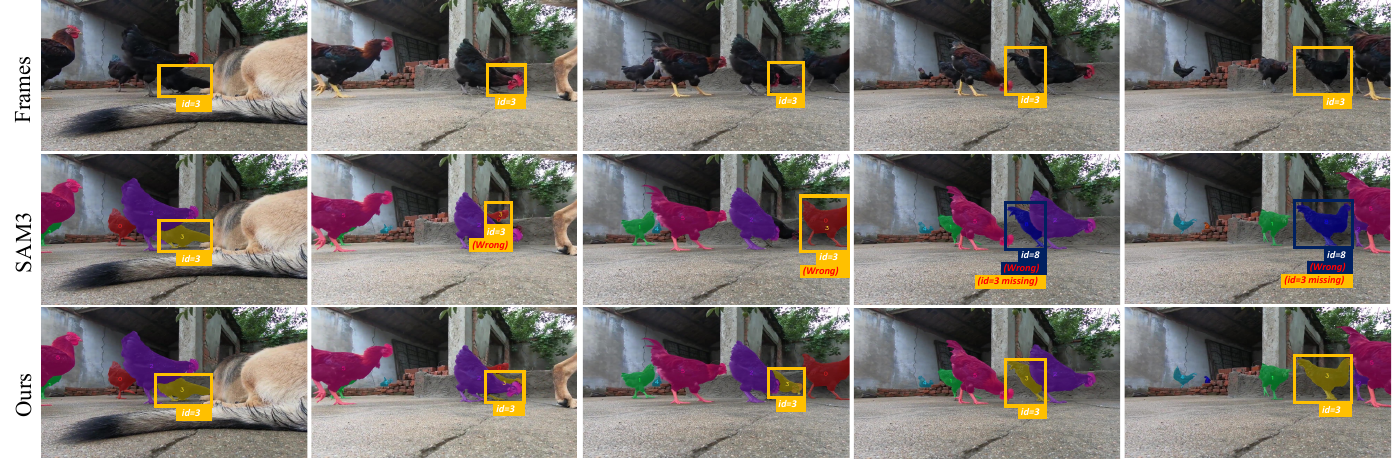}
    \caption{\textit{``The chickens''}}
    \vspace{0.15in}
    \label{fig:case1_a}
  \end{subfigure}
  \begin{subfigure}{\textwidth}
    \centering
    \includegraphics[width=\textwidth]{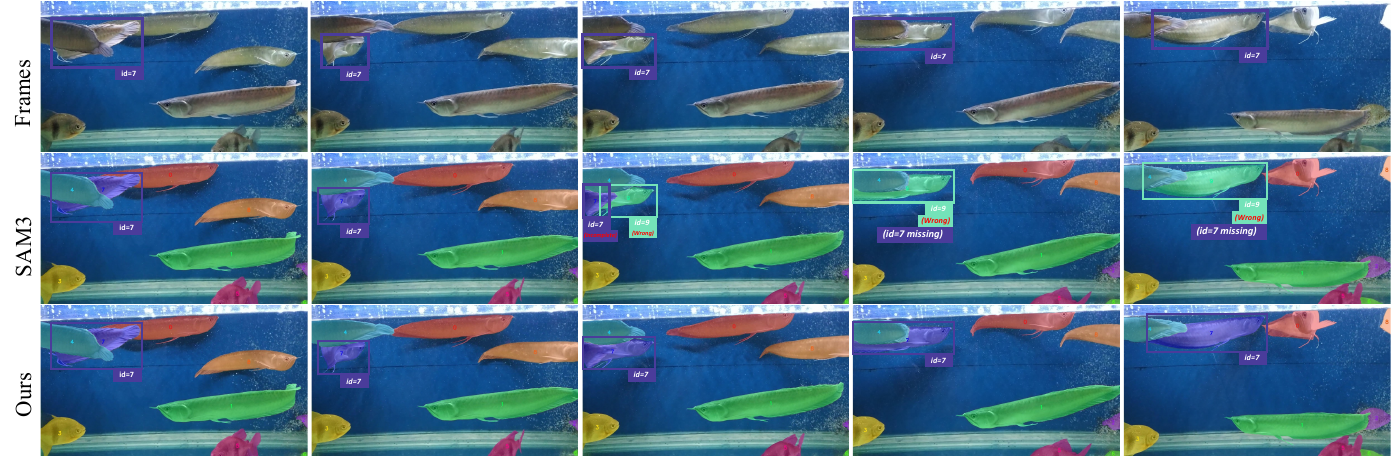}
    \caption{\textit{``The fishes''}}
    \label{fig:case1_b}
  \end{subfigure}
  \caption{Example results of heavy occlusions of targets. (Case 1, zoom in for better view.)}
  \label{fig:case1}
\end{figure*}

\subsection{Qualitative Experiments}
\begin{figure*}[tbp]
  \centering
  \begin{subfigure}{1.0\textwidth}
    \centering
    \includegraphics[width=0.95\textwidth]{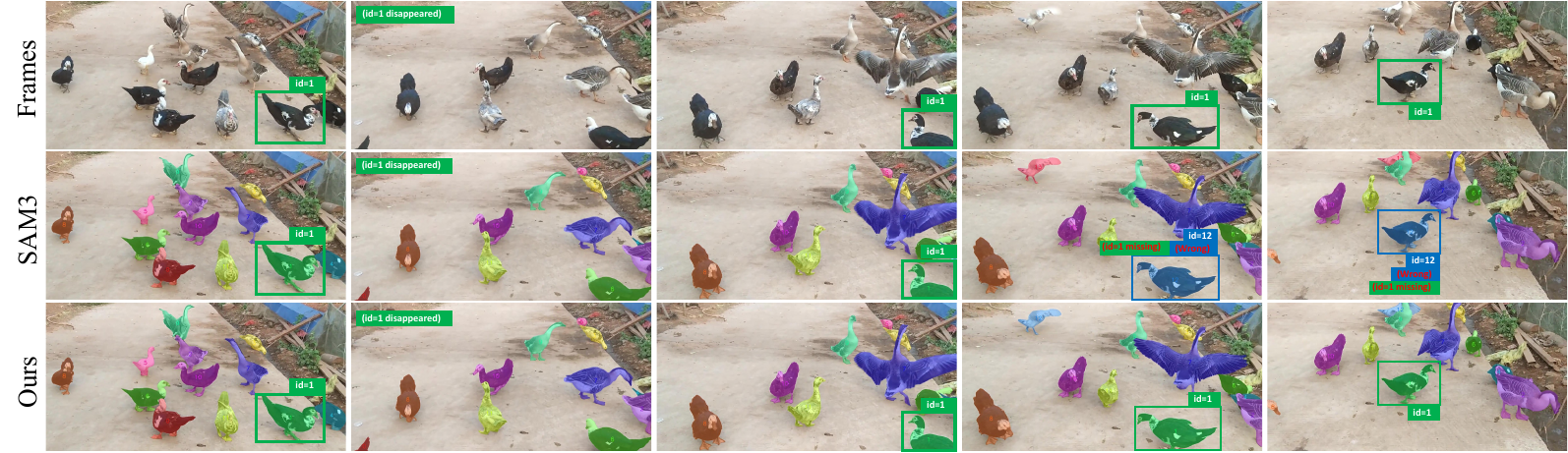}
    \caption{\textit{``The ducks''}}
    \vspace{0.1cm}
    \label{fig:case2_a}
  \end{subfigure}
  
  \begin{subfigure}{1.0\textwidth}
    \centering
    \includegraphics[width=0.95\textwidth]{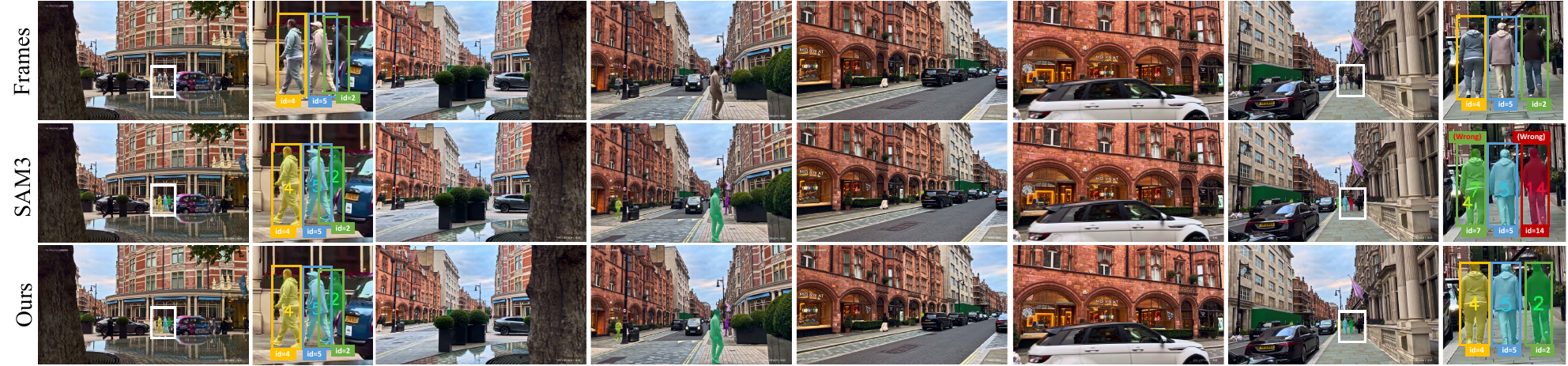}
    \caption{\textit{``Men with jeans''}}
    \vspace{0.1cm}
    \label{fig:case2_b}
  \end{subfigure}

  \begin{subfigure}{1.0\textwidth}
    \centering
    \includegraphics[width=0.95\textwidth]{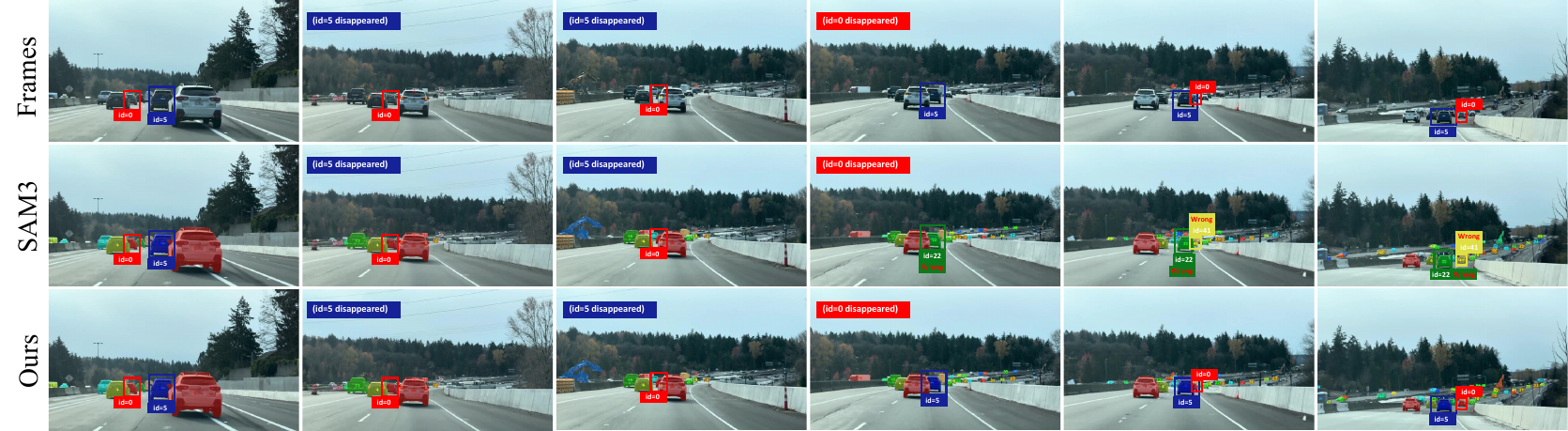}
    \caption{\textit{``Vehicles''}}
  \vspace{-0.05in}
    \label{fig:case2_c}
  \end{subfigure}
  \caption{Example results of disappearance and re-appearance of targets. (Case 2, zoom in for better view.)}
  \vspace{0.12in}
  \label{fig:case2}
\end{figure*}

We present qualitative comparisons on PCS against the SAM3 baseline on challenging scenarios in \cref{fig:case1,fig:case2,fig:case3,fig:case4}.

\noindent \textbf{Case 1: ID switch on heavy occlusion.} \\
As shown in \cref{fig:case1}, when a target is occluded, its visual cues become incomplete, such as chicken neck (top) or fish tail (bottom). These incomplete cues fail to accurately represent the target’s biological features. SAM3 encodes this information into memory based on the average group-level score, leading to subsequent identity drifts. In contrast, SAM3-DMS discards unclear frames, ensuring only clear representations are stored, preventing ID switches.

\noindent \textbf{Case 2}: \textbf{ID switch on disappearance and re-appearance.} \\
As shown in \cref{fig:case2}, when a target exits the view, high confidence from other visible objects might cause SAM3 to sustain a high group average, polluting the target's memory with noisy features, resulting in unrecoverable tracking loss. SAM3-DMS excludes out-of-view frames, ensuring valid and effective object memory for re-identification. We provide additional visualizations in \cref{fig:case2_b} and \ref{fig:case2_c}, featuring scenarios of {multiple targets} exiting and re-appearance,highlighting the robustness of our approach.

\noindent \textbf{Case 3: Interference from distractors.} \\
As illustrated in \cref{fig:case3}, tracking errors frequently occur when distractors exhibit parallel motion as in the corwded fish groups of \cref{fig:case3_a}, or obstruct the target's path as in the pedestrians of \cref{fig:case3_b}. In these scenarios, distractors tend to overshadow the target and significantly suppress its confidence score, thereby preventing memory updates, while SAM3-DMS maintains independent updates for each target, effectively circumventing distractor interference even in very dense and crwoded scene layouts.

\noindent \textbf{Case 4: Random ID drifts.} \\
As shown in \cref{fig:case4}, SAM3 suffers from random and unexpected ID drifts, acting as ``hallucinations" where false positives are generated on totally unrelated objects. This typically occurs under rapid motion, such as the fast-moving plane in \cref{fig:case4_a} or the rapidly walking pedestrian in \cref{fig:case4_b}. Our decoupled approach effectively captures the appearance and motion of each individual target, knowing when a target exits the scene, thereby preventing such drifts.
\section{Discussion and Conclusion}

\textbf{Limitations.} We aim to improve the multi-target video segmentation robustness by improving the memory updating strategy. We enhance the performance for PCS, 
however, for PVS, we mitigate the performance gap between the separate one-by-one inference of each target and simultaneous multi-target inference, but without improving its upper bound. Future research could be conducted to elevate the upper bound of SAM3 for PVS.

\vfill

\noindent \textbf{Conclusion.} In this work, we propose SAM3-DMS, addressing the limitations of SAM3's synchronized memory selection in simultaneous multi-target video segmentation. By introducing a training-free decoupled memory selection strategy, we address the memory pollution issues. Experiments demonstrate that our approach enhances multi-target robustness in various challenging benchmarks, with performance gains scaling positively with target density, establishing a robust foundation for video grounding in the wild.

\begin{figure*}[ht]
  \centering
  \begin{subfigure}{0.95\textwidth}
    \centering
    \includegraphics[width=\textwidth]{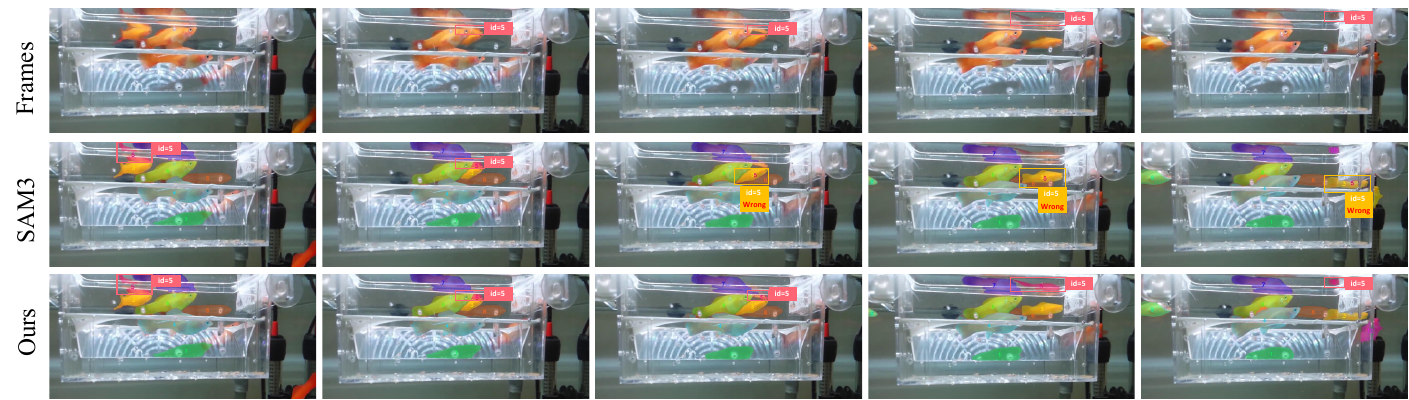}
    \caption{\textit{``The fishes''}}
  \vspace{-0.05in}
    \label{fig:case3_a}
  \end{subfigure}
  \caption{Example results of object interferences, where distractors exhibit parallel motion. (Case 3, zoom in for better view.)}
\end{figure*}%
\begin{figure*}[ht]\ContinuedFloat
  \begin{subfigure}{0.99\textwidth}
    \centering
    \includegraphics[width=\textwidth]{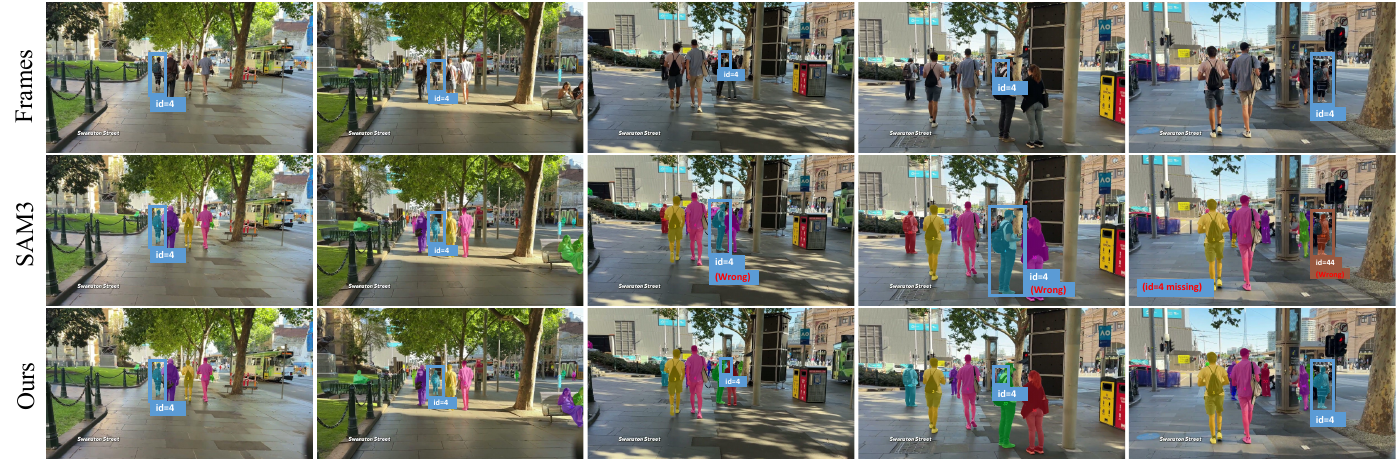}
    \caption{\textit{``Pedestrians''}}
    \label{fig:case3_b}
  \end{subfigure}
  \caption{(cont.) Example results of object interferences, where distractors obstruct the target's path. (Case 3, zoom in for better view.)}
    \vspace{0.3in}
  \label{fig:case3}
\end{figure*}

\begin{figure*}[ht]
  \centering
  \begin{subfigure}{0.99\textwidth}
    \centering
    \includegraphics[width=\textwidth]{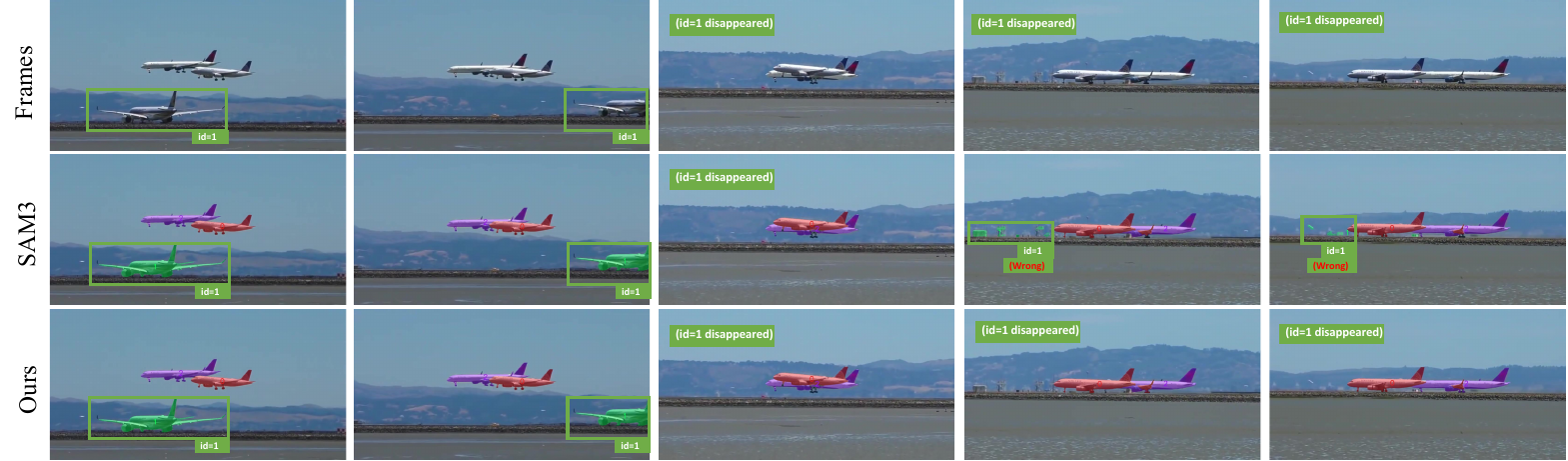}
    \caption{\textit{``The planes''}}
    \vspace{0.15in}
    \label{fig:case4_a}
  \end{subfigure}
  
  \begin{subfigure}{0.99\textwidth}
    \centering
    \includegraphics[width=\textwidth]{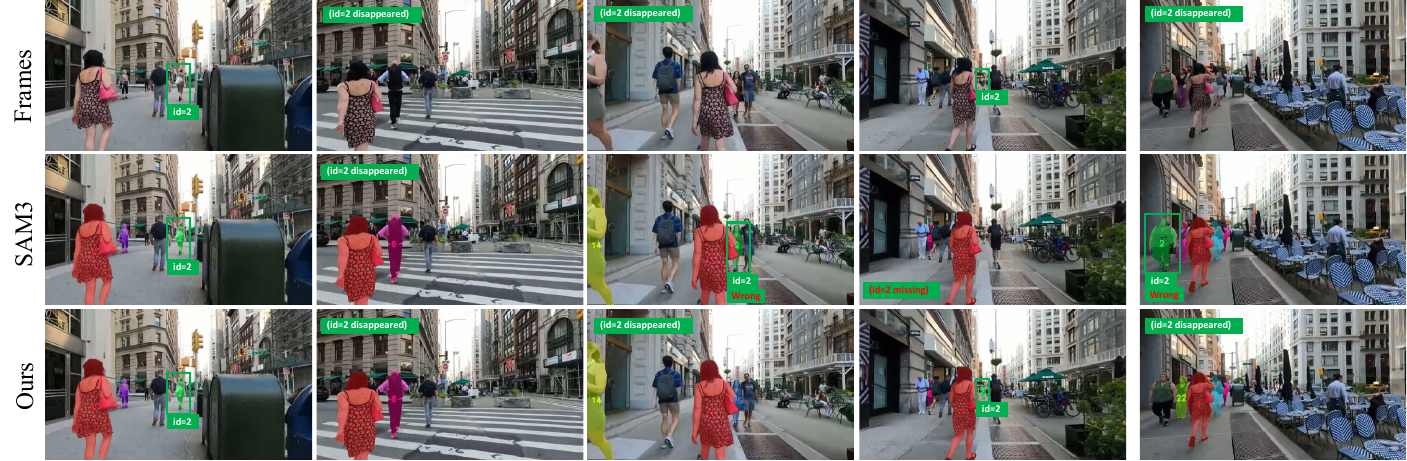}
    \caption{\textit{``Person with handbag''}}
    \vspace{0.1in}
    \label{fig:case4_b}
  \end{subfigure}
  \caption{Example results of rapid motion, which cause SAM3 to exhibit random ID drifts on unexpected objects. (Case 4, zoom in for better view.)}
    \vspace{-0.5in}
  \label{fig:case4}
\end{figure*}